\documentclass[letterpaper,conference]{IEEEtran}
\IEEEoverridecommandlockouts

\usepackage{amsmath,amssymb,amsfonts}
\usepackage{algorithmic}
\usepackage{algorithm}
\usepackage{array}
\usepackage{subcaption}
\usepackage{textcomp}
\usepackage{stfloats}
\usepackage{url}
\usepackage{verbatim}
\usepackage{graphicx}
\usepackage{cite}
\usepackage{pifont}
\usepackage{hyperref}
\usepackage{multirow}
\usepackage{balance}
\usepackage{mathrsfs}

\begin{document}

\title{Towards Validation of Autonomous Vehicles Across Scales using an Integrated Digital Twin Framework}

\author{Tanmay Vilas Samak$^{\ast}$, Chinmay Vilas Samak$^{\ast}$, Venkat Narayan Krovi
\thanks{$^{\ast}$These authors contributed equally.}
\thanks{T. V. Samak, C. V. Samak and V. N. Krovi are with the Automation, Robotics and Mechatronics Laboratory (ARMLab), Department of Automotive Engineering, Clemson University International Center for Automotive Research (CU-ICAR), Greenville, SC 29607, USA. Email: {\tt\small {\{\href{mailto:csamak@clemson.edu}{csamak}, \href{mailto:tsamak@clemson.edu}{tsamak}, \href{mailto:vkrovi@clemson.edu}{vkrovi}\}@clemson.edu}}}}

\maketitle


\begin{abstract}
Autonomous vehicle platforms of varying spatial scales are employed within the research and development spectrum based on space, safety and monetary constraints. However, deploying and validating autonomy algorithms across varying operational scales presents challenges due to scale-specific dynamics, sensor integration complexities, computational constraints, regulatory considerations, environmental variability, interaction with other traffic participants and scalability concerns. In such a milieu, this work focuses on developing a unified framework for modeling and simulating digital twins of autonomous vehicle platforms across different scales and operational design domains (ODDs) to help support the streamlined development and validation of autonomy software stacks. Particularly, this work discusses the development of digital twin representations of 4 autonomous ground vehicles, which span across 3 different scales and target 3 distinct ODDs. We study the adoption of these autonomy-oriented digital twins to deploy a common autonomy software stack with an aim of end-to-end map-based navigation to achieve the ODD-specific objective(s) for each vehicle. Finally, we also discuss the flexibility of the proposed framework to support virtual, hybrid as well as physical testing with seamless sim2real transfer.
\end{abstract}



\section{Introduction}
\label{Section: Introduction}

\IEEEPARstart{T}{he} field of autonomous vehicles has witnessed increasing contributions from a wide spectrum of research and development programs, wherein the choice of underlying autonomous vehicle platform(s) is heavily governed by spatial constraints, safety considerations and cost limitations. However, deploying and validating autonomy algorithms across varying operational scales presents challenges due to factors such as scale-specific dynamics, sensor integration complexities, computational constraints, regulatory considerations, environmental variability, interaction with other traffic participants and scalability concerns, among others. Addressing these challenges is imperative for the seamless integration of autonomy algorithms across different vehicle platforms, which may vary in size and target distinct operating environments.

Digital twins can help alleviate these challenges by providing virtual replicas of the real vehicles and their environments. These autonomy-oriented digital twins, as opposed to conventional simulations, must equally prioritize back-end physics and front-end graphics, which is crucial for the realistic simulation of vehicle dynamics, sensor characteristics and environmental physics. By accurately modeling the interconnect between vehicles, sensors, actuators and the environment, along with traffic participants and infrastructure, digital twins allow for more efficient and cost-effective validation of autonomous systems, thereby reducing the need for extensive real-world testing and accelerating the development process. Additionally, digital twins facilitate iterative design improvements and enable predictive maintenance strategies, ultimately enhancing the safety, reliability and scalability of autonomous vehicle deployments.

\begin{figure}[t]
	\centering
	\includegraphics[width=\linewidth]{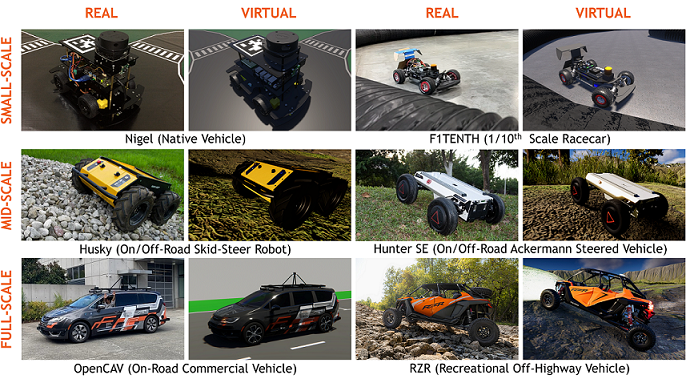}
	\caption{Side-by-side comparison of real and virtual autonomous vehicle platforms spanning across various scales and ODDs: Nigel and F1TENTH (small-scale), Husky and Hunter SE (mid-scale), and OpenCAV and RZR (full-scale).}
	\label{fig1}
\end{figure}

However, in the context of digital twins, seamlessly moving from reality to simulation and back to reality (real2sim2real) requires a streamlined workflow in place. This work proposes AutoDRIVE Ecosystem\footnote{\url{https://autodrive-ecosystem.github.io}} \cite{AutoDRIVEEcosystem, AutoDRIVESimulator, AutoDRIVEReport, AutoDRIVESimulatorReport} as a unified framework for modeling and simulating digital twins of autonomous vehicle platforms across different scales and operational design domains (ODDs), as depicted in Fig. \ref{fig1}. The aim is to streamline the development and validation pipeline of autonomy software stacks, making it agnostic to the physical scale or operating conditions of the underlying vehicle(s). Central to this framework is the development of digital twin representations for a variety of autonomous ground vehicles, spanning multiple scales and targeting distinct ODDs. By leveraging the said digital twins, this paper explores the deployment of a leading open-source autonomy software stack, namely Autoware\footnote{\url{https://autoware.org}} \cite{Autoware}, with an aim of achieving end-to-end map-based navigation tailored to the unique requirements of each vehicle's ODD. Additionally, this paper also discusses the versatility of the proposed framework, emphasizing its capability to support virtual, hybrid and physical testing paradigms \cite{AutoDRIVEMechatronics} while ensuring seamless sim-to-real transfer \cite{AutoDRIVESim2Real2023}. Through this comprehensive approach, the framework aims to facilitate the efficient development, validation and deployment of autonomous vehicle systems across varying scales and operational contexts.


\section{Related Work}
\label{Section: Related Work}

Automotive industry has employed simulators like Ansys Automotive \cite{AnsysAutomotive2021} and Adams Car \cite{AdamsCar2021} to simulate vehicle dynamics at different levels, thereby accelerating the development of its end-products. Since the past few years, however, owing to the increasing popularity of advanced driver-assistance systems (ADAS) and autonomous driving (AD), most of the traditional automotive simulators, such as Ansys Autonomy \cite{AnsysAutonomy2021}, CarSim \cite{CarSim2021} and CarMaker \cite{CarMaker2021}, have started releasing vehicular autonomy features in their updated versions.

Apart from these, several commercial simulators specifically target autonomous driving. These include NVIDIA's Drive Constellation \cite{DRIVEConstellation2021}, Cognata \cite{Cognata2021}, rFpro \cite{rFpro2021}, dSPACE \cite{dSPACE2021} and PreScan \cite{PreScan2021}, to name a few. In the recent past, several research projects have also tried adopting computer games like GTA V \cite{Richter2016, Richter2017, Johnson-Roberson2017} in order to virtually simulate self-driving cars, but they were quickly shut down by the game's publisher.

Finally, the open-source community has also developed several simulators for such applications. Gazebo \cite{Gazebo2004} is a generic robotics simulator natively adopted by Robot Operating System (ROS) \cite{ROS1}. TORCS \cite{TORCS2021}, another open-source simulator widely known in the self-driving community, is probably one of the earliest to specifically target manual and autonomous racing problems. More recent examples include CARLA \cite{CARLA2017}, AirSim \cite{AirSim2018} and Deepdrive \cite{Deepdrive2021} developed using the Unreal \cite{Unreal2021} game engine along with Apollo GameSim \cite{ApolloGameSim2021}, LGSVL Simulator \cite{LGSVLSimulator2020} and AWSIM \cite{AWSIM2023} developed using the Unity \cite{Unity2021} game engine.

The aforementioned simulators pose three key limitations:

\begin{itemize}
    \item Firstly, certain simulation tools prioritize graphical photorealism at the expense of physical accuracy, while others prioritize physical fidelity over graphical realism. In contrast, the AutoDRIVE Simulator achieves a harmonious equilibrium between physics and graphics, offering a variety of configurations to suit diverse computational capabilities.
    \item Secondly, the perception as well as dynamics of varying scales of vehicles and environments differ significantly from each other. Existing simulation tools prefer to target a single vehicle size and ODD. Consequently, transitioning autonomy algorithms from one vehicle platform to the other necessitates considerable additional effort to re-calibrate the autonomy algorithms.
    \item Thirdly, existing simulators may lack precise representations of real-world vehicles or environments, rendering them unsuitable for ``digital twinning'' applications.
\end{itemize}


\section{Methodology}
\label{Section: Methodology}

The core deliverable of this research project was integrating the Autoware Core/Universe stack with AutoDRIVE Ecosystem to demonstrate end-to-end map-based navigation tailored to the unique requirements of 4 different autonomous vehicle platforms, spanning across 3 scales and 3 ODDs. Particularly, we demonstrate small-scale Autoware deployments using Nigel (1:14 scale) \cite{Nigel} and F1TENTH (1:10 scale) \cite{F1TENTH}, two small-scale autonomous vehicle platforms with unique qualities and capabilities. While Nigel targets the autonomous parking ODD, F1TENTH naturally targets the autonomous racing ODD. Mid-scale Autoware deployments are realized using Hunter SE (1:5 scale) \cite{HunterSE}, which target two different ODDs. We employ the Hunter SE to demonstrate autonomous parking in a structured simplistic environment as well as off-road navigation in an unstructured realistic environment. To the best of the authors' knowledge, this is the first-ever off-road deployment of the Autoware stack, thereby expanding its ODD beyond on-road autonomous navigation. Finally, we demonstrate full-scale Autoware deployments using OpenCAV (1:1 scale) \cite{OpenCAV}, which target the autonomous parking ODD in structured simplistic and realistic scenarios.

As a precursor to Autoware deployments, this work discusses the development of vehicle and environment digital twins, which span across different scales and operational design domains. The development of these autonomy-oriented digital twins using AutoDRIVE Ecosystem was, therefore, the primary objective of this research project. This step involved developing geometric as well as dynamics models of vehicles and calibrating them against their real-world counterparts. Additionally, physics-based models for interoceptive as well as exteroceptive sensors and actuators were developed based on their respective datasheets. Finally, creating physically and graphically realistic on-road and off-road environments across scales marked the completion of this objective.

A secondary objective of this research project was to develop cross-platform application programming interfaces (APIs) and human-machine interfaces (HMIs) to connect with AutoDRIVE Ecosystem, which would aid in AutoDRIVE-Autoware integration. This objective, in conjunction with the primary objective, enabled the development of a streamlined real2sim2real framework with deployment demonstrations across varying scales and ODDs.


\section{Digital Twin Framework}
\label{Digital Twin Framework}

The automotive industry has long practiced a gradual transition from virtual, to hybrid, to physical prototyping within an X-in-the-loop (XIL; X = model, software, processor, hardware, vehicle) framework. More recently, digital twins have emerged as potentially viable tools to improve simulation fidelity and to develop adaption/augmentation techniques that can help bridge the sim2real gap. In the following sections, we delve into the development of high-fidelity digital twins of 4 different autonomous vehicles and their operating environments, wherein we also discuss the integration of these with APIs and HMIs for developing autonomy-oriented applications.

\begin{figure*}[t]
	\centering
	\includegraphics[width=\linewidth]{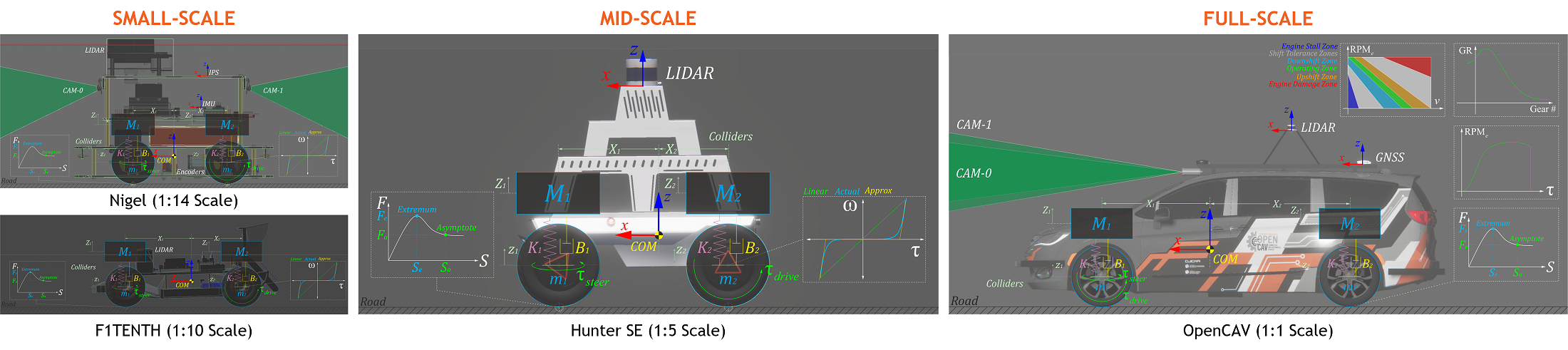}
	\caption{Autonomy-oriented vehicle digital twins across scales and ODDs: Nigel (1:14 scale), F1TENTH (1:10 scale), Hunter SE (1:5 scale), and OpenCAV (1:1 scale) platforms for on/off-road autonomy.}
	\label{fig2}
\end{figure*}

\subsection{Vehicle Models}
\label{Vehicle Models}

The vehicles (refer Fig. \ref{fig2}) are conjunctly modeled using sprung-mass ${^iM}$ and rigid-body representations. Here, the total mass $M=\sum{^iM}$, center of mass, $X_{COM} = \frac{\sum{{^iM}*{^iX}}}{\sum{^iM}}$ and moment of inertia $I_{COM} = \sum{{^iM}*{^iX^2}}$, serve as the linkage between these two representations, where ${^iX}$ represents the coordinates of the sprung masses. Each vehicle's wheels are also modeled as rigid bodies with mass $m$, experiencing gravitational and suspension forces: ${^im} * {^i{\ddot{z}}} + {^iB} * ({^i{\dot{z}}}-{^i{\dot{Z}}}) + {^iK} * ({^i{z}}-{^i{Z}})$.

\subsubsection{Powertrain Dynamics}
\label{Powertrain Dynamics}

For small and mid-scale vehicles, which usually implement an electric motor for propulsion, the front/rear/all wheels are driven by applying a torque ${^i\tau_{drive}} = {^iI_w}*{^i\dot{\omega}_w}$, where ${^iI_w} = \frac{1}{2}*{^im_w}*{^i{r_w}^2}$ represents the moment of inertia, $^i\dot{\omega}_w$ is the angular acceleration, $^im_w$ is the mass, and $^ir_w$ is the radius of the $i$-th wheel. The actuation delays can also be modeled by splitting the torque profile into multiple segments based on operating conditions. For full-scale vehicles, however, the powertrain comprises an engine, transmission and differential. The engine is modeled based on its torque-speed characteristics. The engine RPM is updated smoothly based on its current value $RPM_e$, the idle speed $RPM_i$, average wheel speed $RPM_w$, final drive ratio $FDR$, current gear ratio $GR$, and the vehicle velocity $v$. The update can be expressed as $RPM_e := \left[RPM_i + \left(|RPM_w| * FDR * GR\right)\right]_{(RPM_e,v)}$ where, $[\mathscr{F}]_x$ denotes evaluation of $\mathscr{F}$ at $x$. The total torque generated by the powertrain is computed as $\tau_{\text{total}} = \left[\tau_e\right]_{RPM_e} * \left[GR\right]_{G_\#} * FDR * \tau * \mathscr{A}$. Here, $\tau_e$ is the engine torque, $\tau$ is the throttle input, and $\mathscr{A}$ is a non-linear smoothing operator which increases the vehicle acceleration based on the throttle input. The automatic transmission decides to upshift/downshift the gears based on the transmission map of a given vehicle. This keeps the engine RPM in a good operating range for a given speed: $RPM_e = \frac{{v_{\text{MPH}} * 5280 * 12}}{{60 * 2 * \pi * R_{\text{tire}}}} * FDR * GR$. It is to be noted that while shifting the gears, the total torque produced by the powertrain is set to zero to simulate the clutch disengagement. It is also noteworthy that the auto-transmission is put in neutral gear once the vehicle is in standstill condition and parking gear if handbrakes are engaged in standstill condition. Additionally, switching between drive and reverse gears requires that the vehicle first be in the neutral gear to allow this transition. The total torque $\tau_\text{total}$ from the drivetrain is divided to the wheels based on the drive configuration of the vehicle:
$
\tau_{\text{out}} = \begin{cases}
\frac{\tau_{\text{total}}}{2} & \text{if FWD/RWD} \\
\frac{\tau_{\text{total}}}{4} & \text{if AWD}
\end{cases}
$.
The torque transmitted to wheels $\tau_w$ is modeled by dividing the output torque $\tau_\text{out}$ to the left and right wheels based on the steering input. The left wheel receives a torque amounting to $^{L}\tau_{w} = \tau_{\text{out}} * (1 - \tau_{\text{drop}} * |\delta^{-}|)$, while the right wheel receives a torque equivalent to $^{R}\tau_{w} = \tau_{\text{out}} * (1 - \tau_{\text{drop}} * |\delta^{+}|)$. Here, $\tau_\text{drop}$ is the torque-drop at differential and $\delta^{\pm}$ indicates positive and negative steering angles, respectively. The value of $(\tau_{\text{drop}} * |\delta^{\pm}|)$ is clamped between $[0,0.9]$.

\subsubsection{Brake Dynamics}
\label{Brake Dynamics}

The driving actuators for small and mid-scale vehicles simulate braking torque by applying a holding torque in idle conditions, i.e., ${^i\tau_\text{brake}} = {^i\tau_\text{idle}}$. For full-scale vehicles, the braking torque is modeled as ${^i\tau_\text{brake}} = \frac{{^iM}*v^2}{2*D_\text{brake}}*R_b$ where $R_b$ is the brake disk radius and $D_\text{brake}$ is the braking distance at 60 MPH, which can be obtained from physical vehicle tests. This braking torque is applied to the wheels based on the type of brake input: for combi-brakes, this torque is applied to all the wheels, and for handbrakes, it is applied to the rear wheels only.

\subsubsection{Steering Dynamics}
\label{Steering Dynamics}

The steering mechanism operates by employing a steering actuator, which applies a torque $\tau_{\text{steer}}$ to achieve the desired steering angle $\delta$ with a smooth rate $\dot{\delta}$, without exceeding the steering limits $\pm \delta_\text{lim}$. The rate at which the vehicle steers is governed by its speed $v$ and steering sensitivity $\kappa_\delta$, and is represented as $\dot{\delta} = \kappa_\delta + \kappa_v * \frac{v}{v_\text{max}}$. Here, $\kappa_v$ is the speed-dependency factor of the steering mechanism. Finally, the individual angle for left $\delta_l$ and right $\delta_r$ wheels are governed by the Ackermann steering geometry, considering the wheelbase $l$ and track width $w$ of the vehicle:
$
\left\{
\begin{matrix} 
\delta_l = \textup{tan}^{-1}\left(\frac{2*l*\textup{tan}(\delta)}{2*l+w*\textup{tan}(\delta)}\right) \\ 
\delta_r = \textup{tan}^{-1}\left(\frac{2*l*\textup{tan}(\delta)}{2*l-w*\textup{tan}(\delta)}\right) 
\end{matrix}
\right.
$.

\subsubsection{Suspension Dynamics}
\label{Suspension Dynamics}

For small and mid-scale vehicles, the suspension force acting on each sprung mass is calculated as ${^iM} * {^i{\ddot{Z}}} + {^iB} * ({^i{\dot{Z}}}-{^i{\dot{z}}}) + {^iK} * ({^i{Z}}-{^i{z}})$, where $^iZ$ and $^iz$ denote the displacements of the sprung and unsprung masses, respectively, and $^iB$ and $^iK$ represent the damping and spring coefficients of the $i$-th suspension. For full-scale vehicles, however, the stiffness ${^iK} = {^iM} * {^i\omega_n^2}$ and damping $^iB = 2 * ^i\zeta * \sqrt{{^iK} * {^iM}}$ coefficients of the suspension system are computed based on the sprung mass ${^iM}$, natural frequency ${^i\omega_n}$, and damping ratio ${^i\zeta}$ parameters. The point of suspension force application ${^iZ_F}$ is calculated based on the suspension geometry:
${^iZ_F} = {^iZ_\text{COM}} - {^iZ_w} + {^ir_w} - {^iZ_f}$, where $^iZ_\text{COM}$ denotes the Z-component of vehicle's center of mass, $^iZ_w$ is the Z-component of the relative transformation between each wheel and the vehicle frame ($^VT_{w_i}$), $^ir_w$ is the wheel radius, and $^iZ_f$ is the force offset determined by the suspension geometry. Lastly, the suspension displacement $^iZ_s$ at any given moment can be computed as ${^iZ_s} = \frac{{^iM} * g}{{^iZ_0} * {^iK}}$, where $g$ represents the acceleration due to gravity, and $^iZ_0$ is the suspension's equilibrium point. Additionally, full-scale vehicle models also have a provision to include anti-roll bars, which apply a force on the left ${^LF_r} = K_r * {^RZ} - {^LZ}$ and right ${R^F_r} = K_r * {^LZ} - {^RZ}$ wheels as long as they are grounded at the contact point $Z_c$. This force is directly proportional to the stiffness of the anti-roll bar, $K_r$. The left and right wheel travels are given by ${^LZ} = \frac{-{^LZ_c} - {^Lr_w}}{^LZ_s}$ and ${^RZ} = \frac{-{^RZ_c} - {^Rr_w}}{^RZ_s}$.

\subsubsection{Tire Dynamics}
\label{Tire Dynamics}

Tire forces are determined based on the friction curve for each tire $\left\{\begin{matrix} {^iF_{t_x}} = F(^iS_x) \\{^iF_{t_y}} = F(^iS_y) \\ \end{matrix}\right.$, where $^iS_x$ and $^iS_y$ represent the longitudinal and lateral slips of the $i$-th tire, respectively. The friction curve is approximated using a two-piece spline, defined as $F(S) = \left\{\begin{matrix} f_0(S); \;\; S_0 \leq S < S_e \\ f_1(S); \;\; S_e \leq S < S_a \\ \end{matrix}\right.$, with $f_k(S) = a_k*S^3+b_k*S^2+c_k*S+d_k$ as a cubic polynomial function. The first segment of the spline ranges from zero $(S_0,F_0)$ to an extremum point $(S_e,F_e)$, while the second segment ranges from the extremum point $(S_e, F_e)$ to an asymptote point $(S_a, F_a)$. Tire slip is influenced by factors including tire stiffness $^iC_\alpha$, steering angle $\delta$, wheel speeds $^i\omega$, suspension forces $^iF_s$, and rigid-body momentum ${^iP}={^iM}*{^iv}$. The longitudinal slip $^iS_x$ of $i$-th tire is calculated by comparing the longitudinal components of its surface velocity $v_x$ (i.e., the longitudinal linear velocity of the vehicle) with its angular velocity $^i\omega$: ${^iS_x} = \frac{{^ir}*{^i\omega}-v_x}{v_x}$. The lateral slip $^iS_y$ depends on the tire's slip angle $\alpha$ and is determined by comparing the longitudinal $v_x$ (forward velocity) and lateral $v_y$ (side-slip velocity) components of the vehicle's linear velocity: ${^iS_y} = \tan(\alpha) = \frac{v_y}{\left| v_x \right|}$.

\subsubsection{Aerodynamics}
\label{Aerodynamics}

Small and mid-scale vehicles are modeled with constant coefficients for linear $F_d$ as well as angular $T_d$ drags, which act directly proportional to their linear $v$ and angular $\omega$ velocities. These vehicles do not create significant downforce due to unoptimized aerodynamics, limited velocities and smaller size and mass. Full-scale vehicles, on the other hand, have been modeled to simulate variable air drag $F_\text{aero}$ acting on the vehicle, which is computed based on the vehicle’s operating condition:
$
F_{\text{aero}} = \begin{cases}
F_{d_\text{max}} & \text{if } v \geq v_{\text{max}} \\
F_{d_\text{idle}} & \text{if } \tau_{\text{out}} = 0 \\
F_{d_\text{rev}} & \text{if } (v \geq v_{\text{rev}}) \land (G_\# = -1) \land (RPM_{w} < 0) \\
F_{d_\text{idle}} & \text{otherwise}
\end{cases}
$
where, $v$ is the vehicle velocity, $v_\text{max}$ is the vehicle's designated top-speed, $v_\text{rev}$ is the vehicle's designated maximum reverse velocity, $G_\#$ is the operating gear, and $RPM_w$ is the average wheel RPM. The downforce acting on a full-scale vehicle is modeled proportional to its velocity: $F_\text{down}=K_\text{down}*|v|$, where $K_\text{down}$ is the downforce coefficient.

\subsection{Sensor Models}
\label{Sensor Models}

The simulated vehicles can be equipped with physically accurate interoceptive and exteroceptive sensing modalities.

\subsubsection{Actuator Feedbacks}
\label{Actuator Feedbacks}

Throttle ($\tau$) and steering ($\delta$) sensors are simulated using a simple feedback loop.

\subsubsection{Incremental Encoders}
\label{Incremental Encoders}

Simulated incremental encoders measure wheel rotations $^iN_{\text{ticks}} = {^iPPR} * {^iCGR} * {^iN_{\text{rev}}}$, where $^iN_{\text{ticks}}$ represents the measured ticks, $^iPPR$ is the encoder resolution (pulses per revolution), $^iCGR$ is the cumulative gear ratio, and $^iN_{\text{rev}}$ represents the wheel revolutions.

\subsubsection{Inertial Navigation Systems}
\label{Inertial Navigation Systems}

Positioning systems and inertial measurement units (IMU) are simulated based on temporally coherent rigid-body transform updates of the vehicle $\{v\}$ with respect to the world $\{w\}$: ${^w\mathbf{T}_v} = \left[\begin{array}{c | c} \mathbf{R}_{3 \times 3} & \mathbf{t}_{3 \times 1} \\ \hline \mathbf{0}_{1 \times 3} & 1 \end{array}\right] \in SE(3)$. The positioning systems provide 3-DOF positional coordinates $\{x,y,z\}$ of the vehicle, while the IMU supplies linear accelerations $\{a_x,a_y,a_z\}$, angular velocities $\{\omega_x,\omega_y,\omega_z\}$, and 3-DOF orientation data for the vehicle, either as Euler angles $\{\phi_x,\theta_y,\psi_z\}$ or as a quaternion $\{q_0,q_1,q_2,q_3\}$.

\subsubsection{Planar LIDARs}
\label{Planar LIDARs}

2D LIDAR simulation employs iterative ray-casting \texttt{raycast}\{$^w\mathbf{T}_l$, $\vec{\mathbf{R}}$, $r_{\text{max}}$\} for each angle $\theta \in \left [ \theta_{\text{min}}:\theta_{\text{res}}:\theta_{\text{max}} \right ]$ at a specified update rate. Here, ${^w\mathbf{T}_l} = {^w\mathbf{T}_v} * {^v\mathbf{T}_l} \in SE(3)$ represents the relative transformation of the LIDAR \{$l$\} with respect to the vehicle \{$v$\} and the world \{$w$\}, $\vec{\mathbf{R}} = \left [\cos(\theta) \;\; \sin(\theta) \;\; 0 \right ]^T$ defines the direction vector of each ray-cast $R$, where $r_{\text{min}}$ and $r_{\text{max}}$ denote the minimum and maximum linear ranges, $\theta_{\text{min}}$ and $\theta_{\text{max}}$ denote the minimum and maximum angular ranges, and $\theta_{\text{res}}$ represents the angular resolution of the LIDAR, respectively. The laser scan ranges are determined by checking ray-cast hits and then applying a threshold to the minimum linear range of the LIDAR, calculated as \texttt{ranges[i]}$=\begin{cases} \texttt{hit.dist} & \text{ if } \texttt{ray[i].hit} \text{ and } \texttt{hit.dist} \geq r_{\text{min}} \\ \infty & \text{ otherwise} \end{cases}$, where \texttt{ray.hit} is a Boolean flag indicating whether a ray-cast hits any colliders in the scene, and \texttt{hit.dist}$=\sqrt{(x_{\text{hit}}-x_{\text{ray}})^2 + (y_{\text{hit}}-y_{\text{ray}})^2 + (z_{\text{hit}}-z_{\text{ray}})^2}$ calculates the Euclidean distance from the ray-cast source $\{x_{\text{ray}}, y_{\text{ray}}, z_{\text{ray}}\}$ to the hit point $\{x_{\text{hit}}, y_{\text{hit}}, z_{\text{hit}}\}$.

\subsubsection{Spatial LIDARs}
\label{Spatial LIDARs}

3D LIDAR simulation adopts multi-channel parallel ray-casting \texttt{raycast}\{$^w\mathbf{T}_l$, $\vec{\mathbf{R}}$, $r_{\text{max}}$\} for each angle $\theta \in \left [ \theta_{\text{min}}:\theta_{\text{res}}:\theta_{\text{max}} \right ]$ and each channel $\phi \in \left [ \phi_{\text{min}}:\phi_{\text{res}}:\phi_{\text{max}} \right ]$ at a specified update rate, with GPU acceleration (if available). Here, ${^w\mathbf{T}_l} = {^w\mathbf{T}_v} * {^v\mathbf{T}_l} \in SE(3)$ represents the relative transformation of the LIDAR \{$l$\} with respect to the vehicle \{$v$\} and the world \{$w$\}, $\vec{\mathbf{R}} = \left [\cos(\theta)*\cos(\phi) \;\; \sin(\theta)*\cos(\phi) \;\; -\sin(\phi) \right ]^T$ defines the direction vector of each ray-cast $R$, where $r_{\text{min}}$ and $r_{\text{max}}$ denote the minimum and maximum linear ranges, $\theta_{\text{min}}$ and $\theta_{\text{max}}$ denote the minimum and maximum horizontal angular ranges, $\phi_{\text{min}}$ and $\phi_{\text{max}}$ denote the minimum and maximum vertical angular ranges, and $\theta_{\text{res}}$ and $\phi_{\text{res}}$ represent the horizontal and vertical angular resolutions of the LIDAR, respectively. The thresholded ray-cast hit coordinates $\{x_{\text{hit}}, y_{\text{hit}}, z_{\text{hit}}\}$, from each of the casted rays is encoded into byte arrays based on the LIDAR parameters, and given out as the point cloud data.

\subsubsection{Cameras}
\label{Cameras}

Simulated cameras are parameterized by their focal length $f$, sensor size $\{s_x, s_y\}$, target resolution, as well as the distances to the near $N$ and far $F$ clipping planes. The viewport rendering pipeline for the simulated cameras operates in three stages. First, the camera view matrix $\mathbf{V} \in SE(3)$ is computed by obtaining the relative homogeneous transform of the camera $\{c\}$ with respect to the world $\{w\}$: $\mathbf{V} = \begin{bmatrix} r_{00} & r_{01} & r_{02} & t_{0} \\ r_{10} & r_{11} & r_{12} & t_{1} \\ r_{20} & r_{21} & r_{22} & t_{2} \\ 0 & 0 & 0 & 1 \\ \end{bmatrix}$, where $r_{ij}$ and $t_i$ denote the rotational and translational components, respectively. Next, the camera projection matrix $\mathbf{P} \in \mathbb{R}^{4 \times 4}$ is calculated to project world coordinates into image space coordinates: $\mathbf{P} = \begin{bmatrix} \frac{2*N}{R-L} & 0 & \frac{R+L}{R-L} & 0 \\ 0 & \frac{2*N}{T-B} & \frac{T+B}{T-B} & 0 \\ 0 & 0 & -\frac{F+N}{F-N} & -\frac{2*F*N}{F-N} \\ 0 & 0 & -1 & 0 \\ \end{bmatrix}$, where $L$, $R$, $T$, and $B$ denote the left, right, top, and bottom offsets of the sensor. The camera parameters $\{f,s_x,s_y\}$ are related to the terms of the projection matrix as follows: $f = \frac{2*N}{R-L}$, $a = \frac{s_y}{s_x}$, and $\frac{f}{a} = \frac{2*N}{T-B}$. The perspective projection from the simulated camera's viewport is given as $\mathbf{C} = \mathbf{P}*\mathbf{V}*\mathbf{W}$, where $\mathbf{C} = \left [x_c\;\;y_c\;\;z_c\;\;w_c \right ]^T$ represents image space coordinates, and $\mathbf{W} = \left [x_w\;\;y_w\;\;z_w\;\;w_w \right ]^T$ represents world coordinates. Finally, this camera projection is transformed into normalized device coordinates (NDC) by performing perspective division (i.e., dividing throughout by $w_c$), leading to a viewport projection achieved by scaling and shifting the result and then utilizing the rasterization process of the graphics API (e.g., DirectX for Windows, Metal for macOS, and Vulkan for Linux). Additionally, a post-processing step simulates non-linear lens and film effects, such as lens distortion, depth of field, exposure, ambient occlusion, contact shadows, bloom, motion blur, film grain, chromatic aberration, etc.

\subsection{Digital Twin Calibration}
\label{Digital Twin Calibration}

\begin{figure}[t]
     \centering
     \begin{subfigure}[b]{0.215\linewidth}
         \centering
         \includegraphics[height=1.6cm]{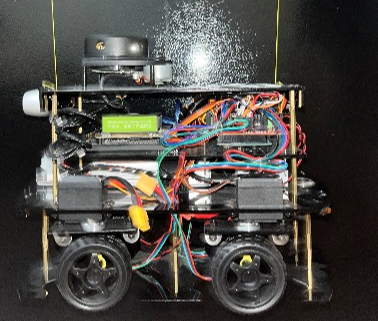}
         \caption{}
         \label{fig3a}
     \end{subfigure}
     \begin{subfigure}[b]{0.24\linewidth}
         \centering
         \includegraphics[height=1.6cm]{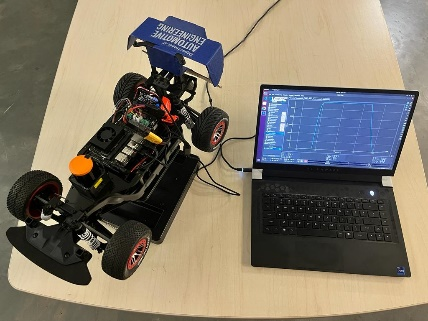}
         \caption{}
         \label{fig3b}
     \end{subfigure}
     \begin{subfigure}[b]{0.24\linewidth}
         \centering
         \includegraphics[height=1.6cm]{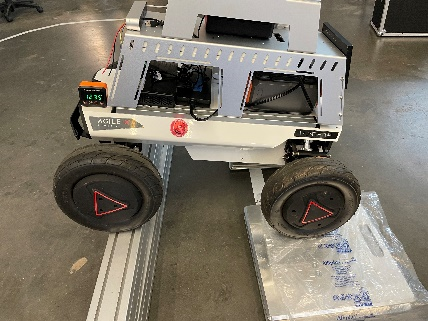}
         \caption{}
         \label{fig3c}
     \end{subfigure}
     \begin{subfigure}[b]{0.24\linewidth}
         \centering
         \includegraphics[height=1.6cm]{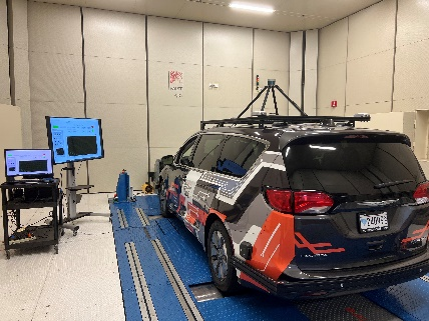}
         \caption{}
         \label{fig3d}
     \end{subfigure}
    \caption{Calibration and validation of vehicle digital twins: (a) System identification of Nigel, (b) VESC calibration of F1TENTH, (c) Static measurements of Hunter SE, and (d) Powertrain measurements of OpenCAV.}
    \label{fig3}
\end{figure}

The vehicle digital twin models were calibrated and validated against geometric, static and dynamic measurement data collected from their real-world counterparts as well as their datasheets. This included the validation of geometric measurements for physical as well as visual purposes, static calibration for mass, center of mass and suspension parameters, and dynamic calibration for validating standard benchmark maneuvers performed in open-loop tests. Additionally, sensor models were validated against static and dynamic characteristics of their real-world counterparts based on their datasheets. Fig. \ref{fig3} depicts some of these calibration/validation tests.

\subsection{Environment Models}
\label{Environment Models}

\begin{figure*}[t]
	\centering
	\includegraphics[width=\linewidth]{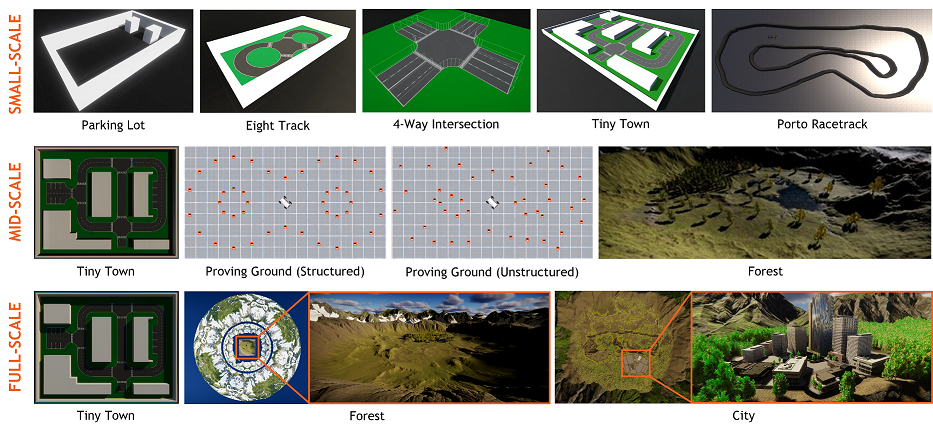}
	\caption{Virtual proving ground environments across different scales and ODDs. Small-scale environments are typically synthetic while mid-scale and full-scale environments can be synthetic as well as realistic.}
	\label{fig4}
\end{figure*}

We leveraged AutoDRIVE Simulator to develop various virtual proving ground environments appropriate for the scales and ODDs of respective host vehicles. Such scenarios can be developed using AutoDRIVE's infrastructure development kit (IDK), Unity's Terrain tools, or other open-standard tools, plugins and assets. Scenarios depicted in Fig. \ref{fig4} include realistic counterparts of small-scale environments such as the Parking Lot, Eight Track, 4-Way Intersection and Tiny Town for Nigel, which were developed using AutoDRIVE IDK, as well as the Porto Racetrack for F1TENTH, which was created based on the binary occupancy grid map of its real-world counterpart. Additionally, simplistic mid-scale and full-scale environments such as the scaled-up versions of Tiny Town along with structured and unstructured Proving Ground scenarios were developed. Finally, two highly detailed mid and full-scale scenarios were developed to support on-road as well as off-road autonomy. These included a City scenario and a Forest environment. The full-scale variants of these scenarios have several rich features and are large enough to support driving for several minutes, if not a few hours. Environmental physics is simulated accurately by conducting mesh-mesh interference detection and computing contact forces, frictional forces, momentum transfer, as well as linear and angular drag acting on all rigid bodies at each time step. Additionally, the simulation of various environmental conditions, such as different times of day as well as weather conditions, can introduce additional degrees of variability.

\subsection{Computational Methods}
\label{Computational Methods}

From a computational perspective, the digital twin framework is developed modularly using object-oriented programming (OOP) constructs. Additionally, the simulator takes advantage of CPU multi-threading as well as GPU instancing (if available) to efficiently handle the workload, while providing cross-platform support. The framework also adopts pre-baked lightmaps, which provide the benefits of physics-based lighting while reducing the computational overhead of real-time raytracing. Furthermore, the simulator implements level-of-detail (LOD) culling to gradually degrade the LOD of environmental objects as they move further away from the scene cameras. However, it is ensured that LOD culling does not affect any of the AV camera sensor(s), thereby striking a balance between computational optimization and simulation fidelity.

\subsection{Digtial Twin Interfaces}
\label{Digtial Twin Interfaces}

The integration of APIs within AutoDRIVE Ecosystem was achieved through the comprehensive expansion and incorporation of AutoDRIVE Devkit. The versatile APIs developed as part of this framework facilitate interactions with the virtual as well as real vehicle platforms and their operating environments using Python, C++, MATLAB/Simulink, ROS\cite{ROS1}, ROS 2 \cite{ROS2}, or the Autoware stack. This expansion caters to a diverse range of programming preferences, empowering users to exploit AutoDRIVE Simulator or AutoDRIVE Testbed for swift and flexible deployment of autonomy algorithms. The framework extends its utility by enabling the development of API-mediated HMIs, catering to both virtual as well as physical vehicles and infrastructure elements.

Furthermore, the simulation framework itself served a dual purpose by not only providing a digital twinning platform, but also enabling the development of direct HMIs to interface with the virtual vehicles and infrastructure. Supported HMI methods to connect with AutoDRIVE Ecosystem include standard keyboard (digital) and mouse (analog), gamepad/joystick (analog) as well as driving and steering rigs (hybrid). This direct-HMI framework, designed for scalability, ensures practical feasibility by relaying identical machine-to-machine (M2M) commands to both virtual and real vehicles as well as infrastructure elements. The versatility of this approach allows for a true digital-twin framework, establishing a seamless connection between the digital space and the physical world. Additionally, in an extended-reality (XR) setup, this framework offers opportunities to extend the direct-HMI teleoperation to real vehicles, enhancing the applicability and potential of AutoDRIVE Ecosystem in diverse operational scenarios.


\begin{figure*}[htpb]
     \centering
     \begin{subfigure}[b]{0.49\linewidth}
         \centering
         \includegraphics[width=\linewidth]{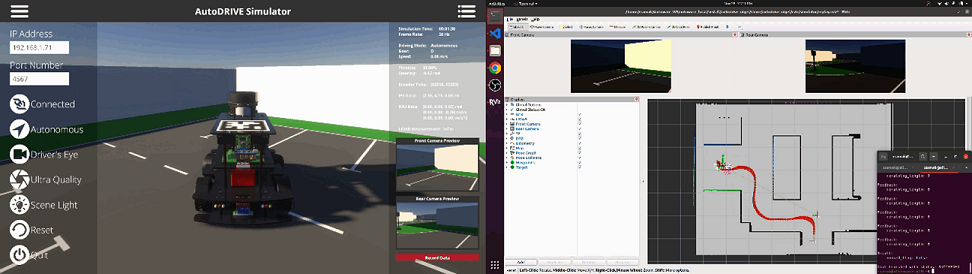}
         \caption{Autonomous parking case study with Nigel in virtual world.}
         \label{fig5a}
     \end{subfigure}
     \begin{subfigure}[b]{0.49\linewidth}
         \centering
         \includegraphics[width=\linewidth]{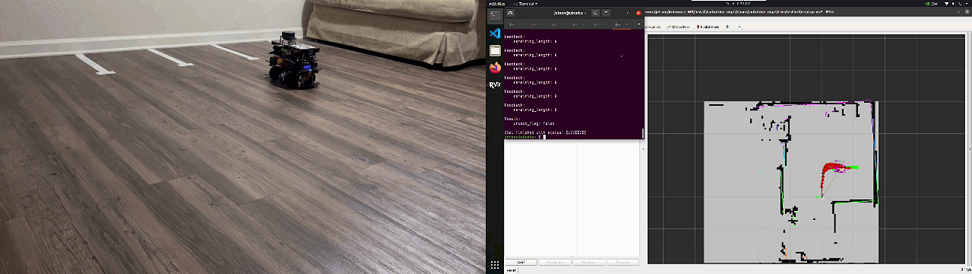}
         \caption{Autonomous parking case study with Nigel in real world.}
         \label{fig5b}
     \end{subfigure}
     \begin{subfigure}[b]{0.49\linewidth}
         \centering
         \includegraphics[width=\linewidth]{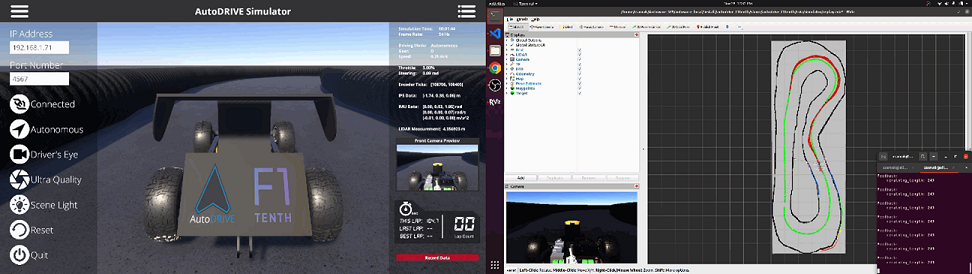}
         \caption{Autonomous racing case study with F1TENTH in virtual world.}
         \label{fig5c}
     \end{subfigure}
     \begin{subfigure}[b]{0.49\linewidth}
         \centering
         \includegraphics[width=\linewidth]{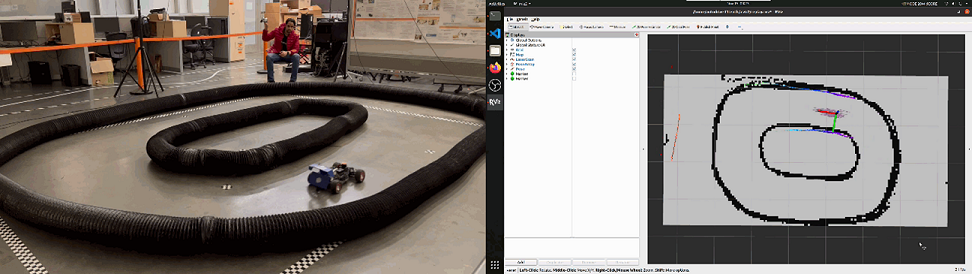}
         \caption{Autonomous racing case study with F1TENTH in real world.}
         \label{fig5d}
     \end{subfigure}
     \begin{subfigure}[b]{0.49\linewidth}
         \centering
         \includegraphics[width=\linewidth]{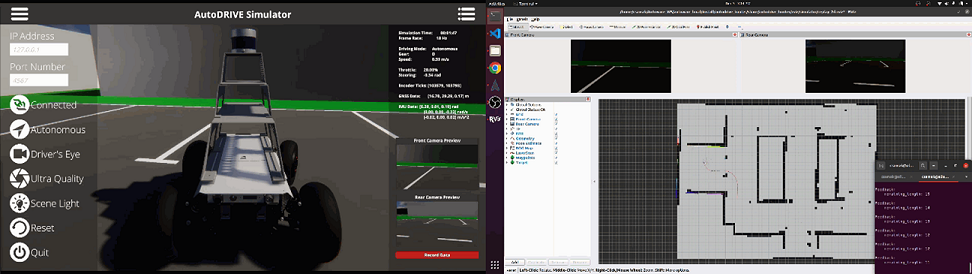}
         \caption{Reduced-order autonomous parking case study with Hunter SE.}
         \label{fig5e}
     \end{subfigure}
     \begin{subfigure}[b]{0.49\linewidth}
         \centering
         \includegraphics[width=\linewidth]{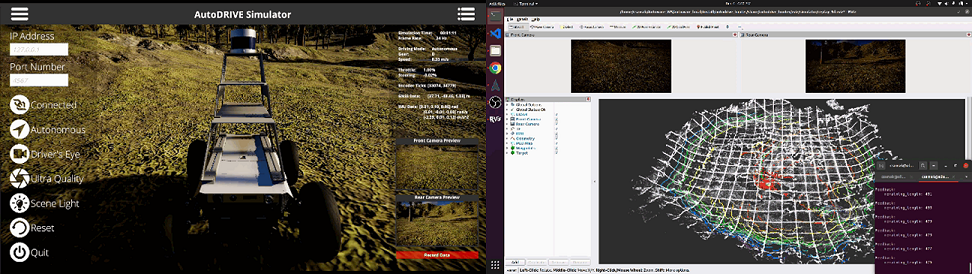}
         \caption{Full-order off-road navigation case study with Hunter SE.}
         \label{fig5f}
     \end{subfigure}
     \begin{subfigure}[b]{0.49\linewidth}
         \centering
         \includegraphics[width=\linewidth]{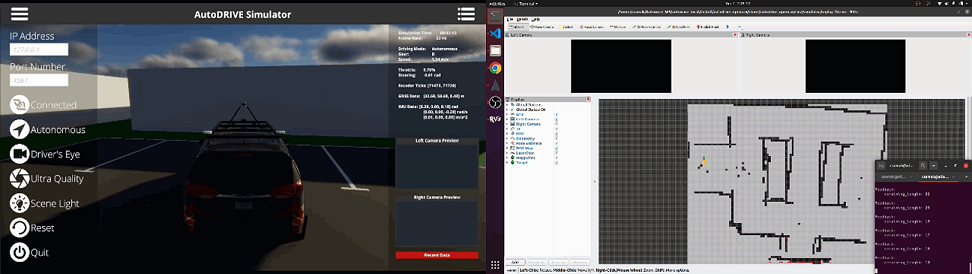}
         \caption{Reduced-order autonomous parking case study with OpenCAV.}
         \label{fig5g}
     \end{subfigure}
     \begin{subfigure}[b]{0.49\linewidth}
         \centering
         \includegraphics[width=\linewidth]{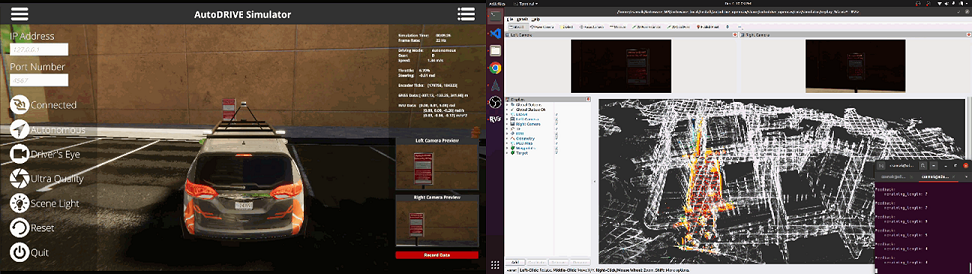}
         \caption{Full-order autonomous parking case study with OpenCAV.}
         \label{fig5h}
     \end{subfigure}
    \caption{Validation of the integrated operation of AutoDRIVE Ecosystem in conjunction with Autoware stack for end-to-end map-based navigation case studies tailored to the unique requirements of each vehicle's scale and ODD. Videos: \texttt{\url{https://youtube.com/playlist?list=PL5Hd4DIMOmEJgpsPYCoLBGNb_91PZVxgA&si=w2wO9h2xKm_IrA1f}}}
    \label{fig5}
\end{figure*}

\section{Case Studies}
\label{Section: Case Studies}

In order to verify the integrated operation of the proposed digital twin framework, namely AutoDRIVE Ecosystem, in conjunction with Autoware, a leading open-source autonomy software stack, a set of experiments were designed in the form of 8 case studies. At a high level, these case studies outline end-to-end map-based navigation tailored to the unique requirements of each vehicle's scale as well as ODD, and work in 3 stages.

\begin{enumerate}
    \item First, the environment is mapped using the LIDAR point cloud data and optionally using odometry estimates by fusing IMU and encoder data while driving (or teleoperating) the vehicle manually.
    \item Next, a reference trajectory is generated by manually driving the vehicle within the (pre)mapped environment, while recording the waypoint coordinates spaced a certain threshold distance apart, using the vehicle's localization estimates with respect to the map's coordinate frame. This can be achieved using just the LIDAR point cloud data, or optionally using odometry estimates by fusing IMU and encoder data. It is worth mentioning that a reference trajectory can also be defined completely offline by using the map information alone, however, in such a case, appropriate measures need to be taken in order to ensure that the resulting trajectory is completely safe and kinodynamically feasible for the vehicle to track in real-time.
    \item Finally, in autonomous mode, the vehicle tracks the reference trajectory using a linearized pure-pursuit controller for lateral motion control and a PID controller for longitudinal motion control.
\end{enumerate}

The exact inputs, outputs, and configurations of perception, planning, and control modules vary with the underlying vehicle platform. Therefore, to keep the overall project\footnote{\url{https://github.com/Tinker-Twins/AutoDRIVE-Autoware}} clean and well-organized, a multitude of custom meta-packages were developed within the Autoware Universe stack to handle different perception, planning, and control algorithms using different input and output information in the form of independent individual packages. Additionally, a separate meta-package was created to handle different vehicles within the AutoDRIVE Ecosystem including Nigel, F1TENTH, Hunter SE and OpenCAV. Each package for a particular vehicle hosts vehicle-specific parameter description configuration files for perception, planning, and control algorithms, map files, RViz configuration files, API program files, teleoperation program files, and user-convenient launch files for getting started quickly and easily. Furthermore, an operational mode is provisioned, which enabled us to transition the small-scale vehicles from simulation to reality, where it is worth mentioning that the exact same controller gains from simulation worked for the real-world deployments.


\section{Conclusion}
\label{Section: Conclusion}

This work investigated the development of autonomy-oriented digital twins of vehicles across different scales and configurations to help support the streamlined deployment of Autoware Core/Universe stack using AutoDRIVE Ecosystem. In essence, this work expands the scope of AutoDRIVE Simulator from catering to scaled autonomous vehicles to developing digital twins of autonomous vehicles across varying scales and ODDs. The core deliverable of this research project was to demonstrate the end-to-end task of map-based navigation. This work discussed the development of vehicle and environment digital twins using AutoDRIVE Ecosystem, along with various application programming interfaces (APIs) and human-machine interfaces (HMIs) to connect with the same. It is worth mentioning that in addition to several Autoware deployment demonstrations, this study described the first-ever off-road deployment of the Autoware stack, thereby expanding its ODD beyond on-road autonomous navigation. In a future work, we seek to investigate multi-agent deployments, dynamic re-planning capabilities, and robust sim2real validation of autonomy algorithms using the proposed framework.


\balance
\bibliographystyle{IEEEtran}
\bibliography{IEEEabrv,References}

\begin{thebibliography}{10}
\providecommand{\url}[1]{#1}
\csname url@samestyle\endcsname
\providecommand{\newblock}{\relax}
\providecommand{\bibinfo}[2]{#2}
\providecommand{\BIBentrySTDinterwordspacing}{\spaceskip=0pt\relax}
\providecommand{\BIBentryALTinterwordstretchfactor}{4}
\providecommand{\BIBentryALTinterwordspacing}{\spaceskip=\fontdimen2\font plus
\BIBentryALTinterwordstretchfactor\fontdimen3\font minus \fontdimen4\font\relax}
\providecommand{\BIBforeignlanguage}[2]{{%
\expandafter\ifx\csname l@#1\endcsname\relax
\typeout{** WARNING: IEEEtran.bst: No hyphenation pattern has been}%
\typeout{** loaded for the language `#1'. Using the pattern for}%
\typeout{** the default language instead.}%
\else
\language=\csname l@#1\endcsname
\fi
#2}}
\providecommand{\BIBdecl}{\relax}
\BIBdecl

\bibitem{AutoDRIVEEcosystem}
\BIBentryALTinterwordspacing
T.~Samak, C.~Samak, S.~Kandhasamy, V.~Krovi, and M.~Xie, ``{AutoDRIVE: A Comprehensive, Flexible and Integrated Digital Twin Ecosystem for Autonomous Driving Research \& Education},'' \emph{Robotics}, vol.~12, no.~3, p.~77, May 2023. [Online]. Available: \url{http://dx.doi.org/10.3390/robotics12030077}
\BIBentrySTDinterwordspacing

\bibitem{AutoDRIVESimulator}
\BIBentryALTinterwordspacing
T.~V. Samak, C.~V. Samak, and M.~Xie, ``{AutoDRIVE Simulator: A Simulator for Scaled Autonomous Vehicle Research and Education},'' in \emph{2021 2nd International Conference on Control, Robotics and Intelligent System}, ser. CCRIS'21.\hskip 1em plus 0.5em minus 0.4em\relax New York, NY, USA: Association for Computing Machinery, 2021, p. 1–5. [Online]. Available: \url{https://doi.org/10.1145/3483845.3483846}
\BIBentrySTDinterwordspacing

\bibitem{AutoDRIVEReport}
\BIBentryALTinterwordspacing
T.~V. Samak and C.~V. Samak, ``{AutoDRIVE - Technical Report},'' 2022. [Online]. Available: \url{https://doi.org/10.48550/arXiv.2211.08475}
\BIBentrySTDinterwordspacing

\bibitem{AutoDRIVESimulatorReport}
\BIBentryALTinterwordspacing
------, ``{AutoDRIVE Simulator - Technical Report},'' 2022. [Online]. Available: \url{https://doi.org/10.48550/arXiv.2211.07022}
\BIBentrySTDinterwordspacing

\bibitem{Autoware}
S.~Kato, S.~Tokunaga, Y.~Maruyama, S.~Maeda, M.~Hirabayashi, Y.~Kitsukawa, A.~Monrroy, T.~Ando, Y.~Fujii, and T.~Azumi, ``{Autoware on Board: Enabling Autonomous Vehicles with Embedded Systems},'' in \emph{2018 ACM/IEEE 9th International Conference on Cyber-Physical Systems (ICCPS)}, 2018, pp. 287--296.

\bibitem{AutoDRIVEMechatronics}
\BIBentryALTinterwordspacing
C.~Samak, T.~Samak, and V.~Krovi, ``{Towards Mechatronics Approach of System Design, Verification and Validation for Autonomous Vehicles},'' in \emph{2023 IEEE/ASME International Conference on Advanced Intelligent Mechatronics (AIM)}, 2023, pp. 1208--1213. [Online]. Available: \url{https://doi.org/10.1109/AIM46323.2023.10196233}
\BIBentrySTDinterwordspacing

\bibitem{AutoDRIVESim2Real2023}
\BIBentryALTinterwordspacing
------, ``{Towards Sim2Real Transfer of Autonomy Algorithms using AutoDRIVE Ecosystem},'' \emph{IFAC-PapersOnLine}, vol.~56, no.~3, pp. 277--282, 2023, 3rd Modeling, Estimation and Control Conference MECC 2023. [Online]. Available: \url{https://www.sciencedirect.com/science/article/pii/S2405896323023704}
\BIBentrySTDinterwordspacing

\bibitem{AnsysAutomotive2021}
\BIBentryALTinterwordspacing
{Ansys Inc.}, ``{Ansys Automotive},'' 2021. [Online]. Available: \url{https://www.ansys.com/solutions/solutions-by-industry/automotive}
\BIBentrySTDinterwordspacing

\bibitem{AdamsCar2021}
\BIBentryALTinterwordspacing
{MSC Software Corporation}, ``{Adams Car},'' 2021. [Online]. Available: \url{https://www.mscsoftware.com/product/adams-car}
\BIBentrySTDinterwordspacing

\bibitem{AnsysAutonomy2021}
\BIBentryALTinterwordspacing
{Ansys Inc.}, ``{Ansys Autonomy},'' 2021. [Online]. Available: \url{https://www.ansys.com/solutions/technology-trends/autonomous-engineering}
\BIBentrySTDinterwordspacing

\bibitem{CarSim2021}
\BIBentryALTinterwordspacing
{Mechanical Simulation Corporation}, ``{CarSim},'' 2021. [Online]. Available: \url{https://www.carsim.com}
\BIBentrySTDinterwordspacing

\bibitem{CarMaker2021}
\BIBentryALTinterwordspacing
{IPG Automotive GmbH}, ``{CarMaker},'' 2021. [Online]. Available: \url{https://ipg-automotive.com/products-services/simulation-software/carmaker}
\BIBentrySTDinterwordspacing

\bibitem{DRIVEConstellation2021}
\BIBentryALTinterwordspacing
{Nvidia Corporation}, ``{NVIDIA DRIVE Sim and DRIVE Constellation},'' 2021. [Online]. Available: \url{https://www.nvidia.com/en-us/self-driving-cars/drive-constellation}
\BIBentrySTDinterwordspacing

\bibitem{Cognata2021}
\BIBentryALTinterwordspacing
{Cognata Ltd.}, ``{Cognata},'' 2021. [Online]. Available: \url{https://www.cognata.com}
\BIBentrySTDinterwordspacing

\bibitem{rFpro2021}
\BIBentryALTinterwordspacing
{rFpro}, ``{Driving Simulation},'' 2021. [Online]. Available: \url{https://www.rfpro.com/driving-simulation}
\BIBentrySTDinterwordspacing

\bibitem{dSPACE2021}
\BIBentryALTinterwordspacing
{dSPACE}, ``{dSPACE},'' 2021. [Online]. Available: \url{https://www.dspace.com/en/pub/home.cfm}
\BIBentrySTDinterwordspacing

\bibitem{PreScan2021}
\BIBentryALTinterwordspacing
{Siemens AG}, ``{PreScan},'' 2021. [Online]. Available: \url{https://tass.plm.automation.siemens.com/prescan}
\BIBentrySTDinterwordspacing

\bibitem{Richter2016}
S.~R. Richter, V.~Vineet, S.~Roth, and V.~Koltun, ``{Playing for Data: Ground Truth from Computer Games},'' in \emph{Proceedings of the European Conference on Computer Vision (ECCV)}, ser. LNCS, J.~Matas, B.~Leibe, M.~Welling, and N.~Sebe, Eds., vol. 9906.\hskip 1em plus 0.5em minus 0.4em\relax Springer International Publishing, 13-15 Nov 2016, pp. 102--118.

\bibitem{Richter2017}
S.~R. Richter, Z.~Hayder, and V.~Koltun, ``{Playing for Benchmarks},'' in \emph{{IEEE} International Conference on Computer Vision, {ICCV} 2017, Venice, Italy, October 22-29, 2017}, 2017, pp. 2232--2241.

\bibitem{Johnson-Roberson2017}
M.~Johnson-Roberson, C.~Barto, R.~Mehta, S.~N. Sridhar, K.~Rosaen, and R.~Vasudevan, ``{Driving in the Matrix: Can Virtual Worlds Replace Human-Generated Annotations for Real World Tasks?}'' in \emph{2017 IEEE International Conference on Robotics and Automation (ICRA)}, 2017, pp. 746--753.

\bibitem{Gazebo2004}
N.~P. Koenig and A.~Howard, ``{Design and use paradigms for Gazebo, an open-source multi-robot simulator},'' in \emph{2004 IEEE/RSJ International Conference on Intelligent Robots and Systems (IROS) (IEEE Cat. No.04CH37566)}, vol.~3, 2004, pp. 2149--2154.

\bibitem{ROS1}
\BIBentryALTinterwordspacing
M.~Quigley, K.~Conley, B.~Gerkey, J.~Faust, T.~Foote, J.~Leibs, R.~Wheeler, and A.~Ng, ``{ROS: An Open-Source Robot Operating System},'' in \emph{ICRA 2009 Workshop on Open Source Software}, vol.~3, Jan 2009. [Online]. Available: \url{http://robotics.stanford.edu/~ang/papers/icraoss09-ROS.pdf}
\BIBentrySTDinterwordspacing

\bibitem{TORCS2021}
\BIBentryALTinterwordspacing
B.~Wymann, E.~Espié, C.~Guionneau, C.~Dimitrakakis, R.~Coulom, and A.~Sumner, ``{TORCS, The Open Racing Car Simulator},'' 2021. [Online]. Available: \url{http://torcs.sourceforge.net}
\BIBentrySTDinterwordspacing

\bibitem{CARLA2017}
A.~Dosovitskiy, G.~Ros, F.~Codevilla, A.~Lopez, and V.~Koltun, ``{CARLA: An Open Urban Driving Simulator},'' in \emph{Proceedings of the 1st Annual Conference on Robot Learning}, ser. Proceedings of Machine Learning Research, S.~Levine, V.~Vanhoucke, and K.~Goldberg, Eds., vol.~78.\hskip 1em plus 0.5em minus 0.4em\relax PMLR, 13-15 Nov 2017, pp. 1--16.

\bibitem{AirSim2018}
S.~Shah, D.~Dey, C.~Lovett, and A.~Kapoor, ``{AirSim: High-Fidelity Visual and Physical Simulation for Autonomous Vehicles},'' in \emph{Field and Service Robotics}, M.~Hutter and R.~Siegwart, Eds.\hskip 1em plus 0.5em minus 0.4em\relax Cham: Springer International Publishing, 2018, pp. 621--635.

\bibitem{Deepdrive2021}
\BIBentryALTinterwordspacing
{Voyage}, ``{Deepdrive},'' 2021. [Online]. Available: \url{https://deepdrive.voyage.auto}
\BIBentrySTDinterwordspacing

\bibitem{Unreal2021}
\BIBentryALTinterwordspacing
{Epic Games Inc.}, ``{Unreal Engine},'' 2021. [Online]. Available: \url{https://www.unrealengine.com}
\BIBentrySTDinterwordspacing

\bibitem{ApolloGameSim2021}
\BIBentryALTinterwordspacing
{Baidu Inc.}, ``{Apollo Game Engine Based Simulator},'' 2021. [Online]. Available: \url{https://developer.apollo.auto/gamesim.html}
\BIBentrySTDinterwordspacing

\bibitem{LGSVLSimulator2020}
G.~Rong, B.~H. Shin, H.~Tabatabaee, Q.~Lu, S.~Lemke, M.~Možeiko, E.~Boise, G.~Uhm, M.~Gerow, S.~Mehta, E.~Agafonov, T.~H. Kim, E.~Sterner, K.~Ushiroda, M.~Reyes, D.~Zelenkovsky, and S.~Kim, ``{LGSVL Simulator: A High Fidelity Simulator for Autonomous Driving},'' in \emph{2020 IEEE 23rd International Conference on Intelligent Transportation Systems (ITSC)}, 2020, pp. 1--6.

\bibitem{AWSIM2023}
\BIBentryALTinterwordspacing
{TIER IV Inc.}, ``{AWSIM},'' 2023. [Online]. Available: \url{https://tier4.github.io/AWSIM}
\BIBentrySTDinterwordspacing

\bibitem{Unity2021}
\BIBentryALTinterwordspacing
{Unity Technologies}, ``{Unity},'' 2021. [Online]. Available: \url{https://unity.com}
\BIBentrySTDinterwordspacing

\bibitem{Nigel}
\BIBentryALTinterwordspacing
C.~V. Samak, T.~V. Samak, J.~M. Velni, and V.~N. Krovi, ``{Nigel -- Mechatronic Design and Robust Sim2Real Control of an Over-Actuated Autonomous Vehicle},'' 2024. [Online]. Available: \url{https://doi.org/10.48550/arXiv.2401.11542}
\BIBentrySTDinterwordspacing

\bibitem{F1TENTH}
\BIBentryALTinterwordspacing
M.~O'Kelly, V.~Sukhil, H.~Abbas, J.~Harkins, C.~Kao, Y.~V. Pant, R.~Mangharam, D.~Agarwal, M.~Behl, P.~Burgio, and M.~Bertogna, ``{F1/10: An Open-Source Autonomous Cyber-Physical Platform},'' 2019. [Online]. Available: \url{https://arxiv.org/abs/1901.08567}
\BIBentrySTDinterwordspacing

\bibitem{HunterSE}
\BIBentryALTinterwordspacing
{AgileX Robotics}, ``{Hunter SE},'' 2023. [Online]. Available: \url{https://global.agilex.ai/chassis/9}
\BIBentrySTDinterwordspacing

\bibitem{OpenCAV}
\BIBentryALTinterwordspacing
{ARMLab CU-ICAR}, ``{OpenCAV: Open Connected and Automated Vehicle},'' 2023. [Online]. Available: \url{https://sites.google.com/view/opencav}
\BIBentrySTDinterwordspacing

\bibitem{ROS2}
\BIBentryALTinterwordspacing
S.~Macenski, T.~Foote, B.~Gerkey, C.~Lalancette, and W.~Woodall, ``{Robot Operating System 2: Design, Architecture, and Uses in the Wild},'' \emph{Science Robotics}, vol.~7, no.~66, p. eabm6074, 2022. [Online]. Available: \url{https://www.science.org/doi/abs/10.1126/scirobotics.abm6074}
\BIBentrySTDinterwordspacing

\end{thebibliography}

\end{document}